%% file: main.tex
\documentclass[letterpaper, 10 pt, journal, twoside]{IEEEtran}
\IEEEoverridecommandlockouts                              % This command is only needed if 
\input{header.tex}
\makeatletter
\IEEEtriggercmd{\reset@font\normalfont\fontsize{8.8pt}{8.8pt}\selectfont}
\makeatother
\IEEEtriggeratref{1}

\title{Learning Neural Traffic Rules}

\author{Xuan Zhang, Xifeng Gao, Kui Wu, Zherong Pan$^\dagger$\\
\vspace{-15px}
\thanks{$^\dagger$ indicates corresponding author. Authors are with Lightspeed Studios, Tencent America. \{xuanxzzhang,xifgao,kwwu,zrpan\}@global.tencent.com.}
}
\IEEEaftertitletext{\vspace{-1.\baselineskip}}
\allowdisplaybreaks
\begin{document}
\maketitle

\newif\ifdetail
\detailfalse

%%%%%%%%%%%%%%%%%%%%%%%%%%%%%%%%%%%%%%%%%%%%%%%%%%%%%%%%%%%%%%%%%%%%%%%%%%%%%%%%
\begin{abstract}
Extensive research has been devoted to the field of multi-agent navigation. Recently, there has been remarkable progress attributed to the emergence of learning-based techniques with substantially elevated intelligence and realism. Nonetheless, prevailing learned models face limitations in terms of scalability and effectiveness, primarily due to their agent-centric nature, i.e., the learned neural policy is individually deployed on each agent.
Inspired by the efficiency observed in real-world traffic networks, we present an environment-centric navigation policy. Our method learns a set of traffic rules to coordinate a vast group of unintelligent agents that possess only basic collision-avoidance capabilities. Our method segments the environment into distinct blocks and parameterizes the traffic rule using a Graph Recurrent Neural Network (GRNN) over the block network. Each GRNN node is trained to modulate the velocities of agents as they traverse through.
Using either Imitation Learning (IL) or Reinforcement Learning (RL) schemes, we demonstrate the efficacy of our neural traffic rules in resolving agent congestion, closely resembling real-world traffic regulations. Our method handles up to $240$ agents at real-time and generalizes across diverse agent and environment configurations.
\end{abstract}

%%%%%%%%%%%%%%%%%%%%%%%%%%%%%%%%%%%%%%%%%%%%%%%%%%%%%%%%%%%%%%%%%%%%%%%%%%%%%%%%
\input{introduction.tex}
\input{related.tex}
\input{problem.tex}
\input{method.tex}
\input{results.tex}
\input{conclusion.tex}
%%%%%%%%%%%%%%%%%%%%%%%%%%%%%%%%%%%%%%%%%%%%%%%%%%%%%%%%%%%%%%%%%%%%%%%%%%%%%%%%
\AtNextBibliography{\footnotesize}
\printbibliography
%\bibliographystyle{IEEEtranS}
%\bibliography{references}
\end{document}

%% file: header.tex
\usepackage{amsmath,amsfonts,amssymb}
\usepackage{prettyref}
\usepackage{mathrsfs}
\usepackage{graphicx}
\usepackage{wrapfig}
\usepackage{subfig}
\usepackage{comment}
\usepackage{MnSymbol}
\usepackage{multirow}
\usepackage{booktabs}
\usepackage{algorithm}
\usepackage[noend]{algpseudocode}
\usepackage[table]{xcolor}
\usepackage{cellspace}
\usepackage
[backend=bibtex,
bibstyle=ieee,
citestyle=numeric,
mincitenames=1,
maxcitenames=2,
natbib=true,
doi=false,
isbn=false,
url=false,
eprint=false]{biblatex}
\usepackage[colorlinks,allcolors=gray,hypertexnames=true]{hyperref} 
\newcommand{\citewithauthor}[1]{\citeauthor{#1} \cite{#1}}
\usepackage{diagbox}

\usepackage{pgf}
\pgfkeys{/pgf/number format/.cd ,precision=2,sci generic={exponent={\times 10^{#1}}}}

\newtheorem{theorem}{Theorem}[section]

\newtheorem{assume}[theorem]{\TE{Assumption}}

\algnewcommand{\LineComment}[1]{\State \(\triangleright\) #1}
\algdef{SE}[DOWHILE]{Do}{doWhile}{\algorithmicdo}[1]{\algorithmicwhile\ #1}
\newcommand*{\colorboxed}{}
\def\colorboxed#1#{%
  \colorboxedAux{#1}%
}
\newcommand*{\colorboxedAux}[3]{%
  \begingroup
    \colorlet{cb@saved}{.}%
    \color#1{#2}%
    \boxed{%
      \color{cb@saved}%
      #3%
    }%
  \endgroup
}
\newrefformat{fig}{Figure~\ref{#1}}
\newrefformat{par}{Section~\ref{#1}}
\newrefformat{appen}{Appendix~\ref{#1}}
\newrefformat{sec}{Section~\ref{#1}}
\newrefformat{sub}{Section~\ref{#1}}
\newrefformat{table}{Table~\ref{#1}}
\newrefformat{ass}{Assumption~\ref{#1}}
\newrefformat{alg}{Algorithm~\ref{#1}}
\newrefformat{def}{Definition~\ref{#1}}
\newrefformat{thm}{Theorem~\ref{#1}}
\newrefformat{lem}{Lemma~\ref{#1}}
\newrefformat{step}{Step~\ref{#1}}
\newrefformat{ln}{Line~\ref{#1}}
\newrefformat{eq}{Equation~\ref{#1}}
\newrefformat{pb}{Problem~\ref{#1}}
\newrefformat{it}{Item~\ref{#1}}
\newrefformat{te}{Term~\ref{#1}}
\def\Eqref Eq:#1:{\eqref{eq:#1}}
\newrefformat{Eq}{Equation~\Eqref#1:}

\newcommand{\TE}[1]{\textbf{#1}}

\newcommand{\argmin}[1]{\underset{#1}{\text{argmin}}}
\newcommand{\argminP}[1]{\text{argmin}}
\newcommand{\argmax}[1]{\underset{#1}{\text{argmax}}}
\newcommand{\argmaxP}[1]{\text{argmax}}
\newcommand{\ST}{\text{s.t.}}

\newcommand{\HH}{\text{H}}
\newcommand{\SP}{\text{SP}}
\newcommand{\EConv}{\text{EConv}}
\newcommand{\ID}{\text{I}}
\newcommand{\TRU}{\text{RFU}}
\newcommand{\MAX}{\text{max}}
\newcommand{\SEED}{\text{S}}
\newcommand{\DELETE}{\text{D}}
\newcommand{\FREE}{\text{F}}
\newcommand{\SD}{\text{SD}}
\newcommand{\FOC}{\text{F}}
\newcommand{\NAVI}{\text{LNavi}}
\newcommand{\proofread}[1]{}
\newif\ifArxiv
\usepackage{xcolor}
\definecolor{Blue}{rgb}{0.2, 0.2, 0.8}
\definecolor{Black}{rgb}{0.0, 0.0, 0.0}
\newcommand{\revise}[1]{\textcolor{Black}{#1}}

%% file: introduction.tex
\section{Introduction} \label{sec:introduction}
Multi-agent navigation forms the cornerstone of numerous pivotal robotic applications, including domains such as automated warehousing~\cite{azadeh2019robotized}, autonomous driving~\cite{yurtsever2020survey}, and realization of smart cities~\cite{silva2018towards}. As a consequence, the refinement of navigation algorithms has garnered substantial research focus throughout the past decades. An ideal navigation algorithm should satisfy three desiderata. Scalability: Capable of controlling an arbitrarily large crowd of agents; Generality: Capable of handling arbitrary environment and agent configurations; Efficacy: A mild growth of computational and deployment cost with crowd size. Unfortunately, to this day, an ideal navigation algorithm that satisfies all three attributes continues to be elusive.

The key to a successful navigation algorithm is the strategy to mitigate agent congestion. Early research endeavors tackled congestion through either back-tracking search~\cite{de2013push,yu2013structure,sharon2015conflict} or localized collision-avoidance~\cite{van2008reciprocal,karamouzas2017implicit}. The search-based algorithms exhibit notable generality; some even provide assurances of completeness or optimality. However, they have limitations with scalability and efficacy. Conversely, the local navigation policies exhibit ideal scalability, yet their generality is curtailed due to their myopic nature, oftentimes leading to agents stuck in congested configurations. Recent strides in this domain have been achieved through learning-based navigation policies~\cite{fan2020distributed,ji2021decentralized,puente2022review}. These innovative algorithms parameterize navigation policies as deep networks, subsequently fine-tuned through either IL or RL to address complex, long-term decision-making challenges. While learning-based policies empirically surpass myopic local navigation methods, they necessitate that each agent possesses the computational resources to execute intricate neural network inferences. This assumption results in a substantial deployment cost. In reality, large-scale robot swarm systems~\cite{wei2010sambot,rubenstein2012kilobot} are often equipped with limited computational capabilities that cannot feasibly accommodate the demands of neural network inferences.
\begin{comment}
\begin{figure}[t]
\centering
\includegraphics[width=.99\linewidth]{figs/real-world.jpg}
\caption{\label{fig:traffic}An environment-centric real-world traffic system where agents exhibit minimal intelligence and avoid congestion by following the traffic rules.}
\vspace{-10px}
\end{figure}
\end{comment}

Examining prevailing learning-based policies~\cite{fan2020distributed,li2020graph,ji2021decentralized}, we find a vast majority of them being agent-centric, i.e., their deep neural policies are independently deployed on each individual agent, \revise{which inevitably requires non-trivial computation and communication capabilities.} In contrast, real-world traffic networks exhibit an environment-centric approach rather than an agent-centric one. At the heart of these networks resides a collection of human-designed traffic rules driven solely by the specific characteristics of the environment, such as highways, crossroads, and T-junctions. The efficacy of these rules is evidenced in their ability to manage substantial crowds of pedestrians and vehicles, effortlessly mitigating congestion by adhering to predefined traffic regulations. In essence, this approach sidesteps the need for intricate agent-specific decision-making processes.

\TE{Main Results:} Inspired by the elegance of real-world traffic networks, we introduce a learnable environment-centric navigation policy, assuming that our agents are endowed solely with rudimentary computational capacities, confining their interactions to local navigations. Our neural policy encapsulates a set of traffic rules based on the environments, which are followed by our agents. Specifically, our method decomposes the environment into discrete blocks, and we represent our neural policy as a GRNN built on the block network. Each GRNN node is trained to manipulate the velocities of agents traversing the corresponding block, facilitating the rule-following behavior. Our GRNN is trained to mitigate agent congestion via two methodologies: IL, guided by a groundtruth traffic rule, and RL, guided by a congestion-resolving reward signal. After the training phase, our neural policy can be deployed on unseen environment configurations to coordinate up to $240$ intelligent agents at real-time in a simulated environment. \revise{Our contributions are summarized below:}
\begin{itemize}
\item \revise{A decentralized agent navigation paradigm utilizing learned environment-encoded traffic rules.}
\item \revise{Environment-centric navigation policies parameterization using GRNN.}
\item \revise{Reward design and training algorithms for environment-centric policies in both IL and RL setting.}
\end{itemize}

%% file: related.tex
\section{\label{sec:related}Related Work}
We review related works on multi-agent navigation, learning-based navigation methods, and traffic-rule modeling.

\TE{Multi-Agent Navigation:} Over the course of time, researchers have branched into two avenues, yielding centralized and decentralized navigation algorithms. Centralized algorithms operate under the premise that a central computational node coordinates the movements of all agents. This node employs various search-based algorithms to avert congestion over a long horizon. One noteworthy example of this paradigm is the conflict-based search algorithm~\cite{sharon2015conflict}. It is a variant of the branch-and-bound approach, designed to resolve motion conflicts among agents iteratively converging to the optimal navigation plan. While such methods theoretically accommodate any environment or agent configuration, their practical computational demands surge with the number of agents, as analyzed in~\cite{yu2013structure}. Consequently, real-world centralized algorithms either compromise optimality for efficiency~\cite{yu2020average} or lean on additional assumptions~\cite{solovey2015motion}. Despite their comprehensive applicability, centralized algorithms exert a substantial communication cost, necessitating synchronization among all agents in each step. 

Considering their limitations, research endeavors have veered towards decentralized algorithms. In this context, each agent autonomously plans and executes its motion asynchronously. Prominent decentralized techniques~\cite{van2008reciprocal, karamouzas2017implicit} only mandate agents to engage in rudimentary collision avoidance with local neighbors, devoid of extensive long-horizon decision-making. Astonishingly, these straightforward algorithms capably resolve congestion for sizable agent swarms in environments replete with expansive open spaces. Nonetheless, in intricate settings with narrow passages, agents can become ensnared. Such deadlock configurations can be alleviated by introducing heuristic congestion-resolving behaviors~\cite{yang2020review}, such as grouping~\cite{he2016dynamic}, yielding~\cite{10160902}, and local coordination~\cite{schuerman2010situation}. Regrettably, while these heuristic approaches ameliorate specific types of deadlock scenarios, none possess the potential to address all scenarios comprehensively. In contrast, our proposed method presents a cohesive strategy for tackling congestion issues while preserving the efficiency and scalability inherent in decentralized algorithms. Our neural policy embeds environment-centric traffic regulations and operates under the assumption of minimal computational capabilities on the agents' part.

\TE{Learning-based Navigation Methods:} The incorporation of data-driven techniques has proven highly effective in enhancing the performance of navigation algorithms. In the realm of centralized algorithms, \citewithauthor{huang2021learning} introduced a method to learn a conflict selection policy, thereby expediting the conflict-based search process. Another significant advancement comes from the work of \citewithauthor{HanYu20RAL}, who harnessed a database of small-scale navigation plans to accelerate the search for larger-scale counterparts. Most other data-driven approaches are designed to work with decentralized algorithms. The challenge of addressing deadlock concerns in local navigation~\cite{van2008reciprocal, karamouzas2017implicit} prompted \citewithauthor{fan2020distributed} to propose a neural navigation policy, trained through RL, for mitigating long horizon congestion. However, their neural policy's reliance solely on local environmental information precludes it from surpassing the performance of conventional local navigation algorithms.

Enhancements to neural policies have subsequently emerged, including the use of global environment maps~\cite{tan2020deepmnavigate} and the facilitation of inter-agent communication~\cite{li2020graph, blumenkamp2021emergence}. Notably, these approaches, driven by an agent-centric philosophy, mandate each agent to possess the non-trivial capability of network inference. In this manner, the more sophisticated network architectures of~\cite{tan2020deepmnavigate,li2020graph,blumenkamp2021emergence} inevitably introduce a heavier computational overhead.

On another front, we are aware of two recent endeavors~\cite{ji2021decentralized, ye2023differentiable} that bear resemblance to our partially environment-centric approach. \citewithauthor{ji2021decentralized} introduced an augmented GNN policy, introducing environmental router nodes in addition to agent nodes. However, agents remain modeled as GNN nodes, necessitating agent-wise network inferences. On the other hand, \citet{ye2023differentiable} proposed a kernel-based neural policy wherein a small set of kernels dictates all agents' velocities. Nevertheless, this approach mandates agents to operate in a centralized manner, incurring substantial communication costs anew. In contrast, our method adheres solely to an environment-centric framework, demanding only rudimentary computational capabilities from the agents. Moreover, our method operates in a fully decentralized manner, eliminating the need for agent coordination and requiring them solely to consult the environment-centric GRNN policy for velocity modulation.

\TE{Traffic-Rule Modeling:} The remarkable versatility, scalability, and efficiency exhibited by real-world traffic networks have spurred a substantial volume of research aimed at analyzing, emulating, and generalizing these networks to novel contexts. One avenue of inquiry revolves around data-driven traffic simulations and predictions~\cite{chao2017realistic, ma2019trafficpredict, suo2021trafficsim}. Guided by real-world agent trajectory datasets, these methods train a probabilistic model to forecast agents' forthcoming trajectories. However, these probabilistic models remain oblivious to the inherent physical dynamics governing agents, often failing to ensure collision-free interactions among them. A complementary line of work leverages RL to enhance real-world traffic networks through traffic light control strategies~\cite{yau2017survey}.

Amidst these environment-centric approaches, traffic rules play an important role in agent-centric autonomous driving algorithms to ensure driving safety. For instance, \citewithauthor{greenhalgh2014recognizing} introduced a model for text-based traffic sign recognition, while \citewithauthor{franklin2020traffic} devised a framework to detect violations of traffic signals. Additionally, \citewithauthor{lee2003real} formulated a method for predicting imminent freeway collisions. When compared with our method, it becomes evident that all the aforementioned techniques are heavily engineered toward the specific set of real-world traffic rules. In contrast, we advocate the utilization of a neural policy to encode an arbitrary set of unknown traffic rules, which can be learned to resolve congestion for arbitrary environmental and agent contexts. This dynamic approach not only engenders adaptability, but also accommodates scenarios where conventional, pre-defined rules may not suffice.

%% file: problem.tex
\input{pipeline.tex}
\section{\label{sec:problem}Problem Statement}
In this section, we present the formulation of the multi-agent navigation problem, addressing a scenario involving a swarm of agents operating in a 2D environment and we take the following standard assumption:
\begin{assume}
\revise{All agents are circular-shaped with radius $r$. Agents are omnidirectional and capable of instantaneous change of moving direction. Agents move in a rectangular static environment with arbitrary interior obstacles, also of rectangular shapes.}
\end{assume}
The position of the $i$th agent at the $t$-th time step is denoted as $x_i^t\in\mathbb{R}^2$. The principal objective of each agent is to traverse from its initial position $x_i^0$ to a predefined goal location $g_i$, while ensuring a trajectory devoid of collisions. Our approach is built off of the local navigation algorithms~\cite{van2008reciprocal,karamouzas2017implicit}, which offer a guarantee of collision-free motion. These algorithms take as input the desired velocity $v_i^t$ for each agent at each time step, and then output the positions of the agents in the subsequent time step by approximating the solution to the following optimization problem:
\begin{equation}
\begin{aligned}
\label{eq:simulator}
\argmin{x_i^{t+1}}\;&\sum_i\|x_i^{t+1}-x_i^t-v_i^t\|^2\\
\ST\quad&\|x_i^{t+\gamma}-x_j^{t+\gamma}\|\geq 2r\quad\forall i\neq j\land\gamma\in[0,1],
\end{aligned}
\end{equation}
where $x_i^{t+\gamma}$ denotes the interpolated position of an agent at a fractional time instance $t+\gamma$. This optimization moves agents via the desired velocity as much as possible, while the associated constraints ensure a minimum separation distance of $2r$ is maintained at every time instance. 

Notably, local navigation algorithms are limited to ensuring collision avoidance, leaving the high-level behavior of the agents to be specified by the desired velocities $v_i^t$. The primary objective of our work lies in developing a neural navigation policy to predict $v_i^t$ for all agents to facilitate navigation tasks while minimizing instances of congestion. In addition to the settings outlined above, our study incorporates two key assumptions, considering the constraints posed by existing swarm robot hardware~\cite{wei2010sambot,rubenstein2012kilobot}. First, we operate under the assumption that agents possess limited computational resources. Specifically, our requirement is solely for the agents to compute a local collision-free velocity, and no other substantial computations are permissible. Second, we consider the agents to possess restricted communication capabilities. In this context, agents are permitted to independently and asynchronously gather information from proximate computational nodes, while coordination of motion between any two agents is not allowed.

%% file: pipeline.tex
\begin{figure*}[t]
\centering
\includegraphics[width=.99\linewidth]{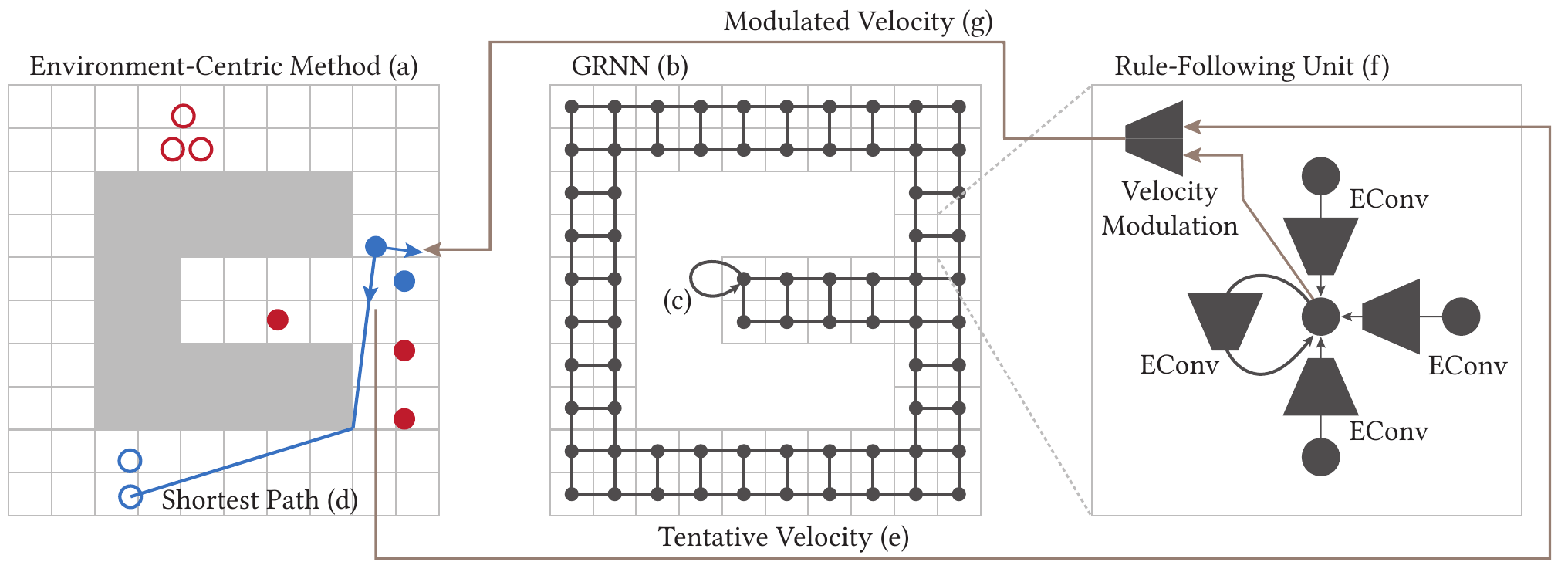}
\caption{\label{fig:pipeline}Two distinct groups of agents (red and blue circles) are engaged in navigation, each with its designated goal positions (red or blue hollow circles). Our environment-centric technique (a) divides the environment into discrete blocks, upon which a GRNN (b) is deployed based on the block network. The RGNN inference process is executed just once, preceding the commencement of the navigation task, where the recurrent units update the internal state of each block and encode the pertinent traffic rules (c). 
During each timestep, an agent calculates the shortest path (d) to yield a tentative velocity (e). The agent then indexes the block it resides within, and proceeds to query the block's rule-following unit (f), resulting in a congestion-resolving modulated velocity (g).}
\end{figure*}

%% file: method.tex
\section{\label{sec:method}Neural Traffic Rule Model}
Our approach is inspired by the inherent resemblance between real-world traffic networks and the specific challenge outlined in~\prettyref{sec:problem}. Real-world agents execute merely basic localized collision resolutions, steering clear of close-by pedestrians or the vehicle immediately ahead. Additionally, the entire set of agents navigates with minimal coordination. Our neural navigation policy is designed to mimic this paradigm. We optimize a parameterized set of traffic rules (\prettyref{sec:rule}) that are consistently followed by all agents in a decentralized manner (\prettyref{sec:agent}).

\subsection{\label{sec:rule}Neural Traffic Rule Parameterization}
The comprehensive framework of our methodology is illustrated in \prettyref{fig:pipeline}. We begin by recognizing that traffic rules are dictated solely by the local characteristics of the environment. For instance, the rules governing a crossroad remain unaffected by the type of road situated a block away. We incorporate this principle by segmenting the free space into square blocks. This arrangement constructs a block network, where each block maintains connections with its adjacent neighbors. To facilitate bidirectional traffic within each passage, we ensure that our block network has a minimum width of 2 blocks (\prettyref{fig:pipeline}b). Furthermore, a noteworthy feature is that traffic rules exhibit independence from individual agents. This attribute empowers a single set of rules to address arbitrary agent navigation scenarios without solving intricate decision-making problems. In view of this, we parameterize these agent-agnostic traffic rules via a GRNN as introduced by~\citewithauthor{ruiz2020gated}.

The core of our GRNN resides in the hidden-state representation, denoted as $h_{ij}$, for the $ij$th block. This representation is continually updated through an Edge Convolution (\EConv) mechanism, iteratively realized as follows (\prettyref{fig:pipeline}f):
\begin{align}
h_{ij}^{k+1}=\HH(h_{ij}^{k},\oplus_{kl\in\mathcal{N}{ij}}\EConv(x_{ij}-x_{kl},h_{kl}^k)).
\end{align}
In this context, $x_{ij}$ denotes the spatial coordinates of the center of the $ij$th block, and $\mathcal{N}_{ij}$ represents its set of neighboring blocks. The function $\EConv(\bullet)$ is parameterized as a Multi-Layer Perceptron (MLP) that encodes the information from each neighboring block. The cumulative information propagation is subsequently summarized via the $\oplus$ operator. We have experimented with $\oplus$ being either concatenation or summation and the results are satisfactory in both cases. This aggregated information is then channeled through the hidden state update function $\HH(\bullet)$, itself parameterized by another MLP. Starting with $h_{ij}^0=0$, the GRNN progressively refines all $h_{ij}^k$ until a fixed point $h_{ij}^k\to h_{ij}^\star$ is attained. The converged states implicitly encapsulate the essential traffic rules. In practical application, we approximate the convergence by performing $K$ updates.

Since our GRNN is completely independent of agents, a single inference step is adequate to derive $h_{ij}^K$ prior to the commencement of navigation tasks, ensuring runtime efficiency. Notably, various design approaches for our traffic rule network exist, and interested readers can refer to~\cite{hamilton2017inductive} for an overview of design insights. The main objective of our GRNN framework is to learn traffic rules from one environment and generalize them to different environments with distinct topologies in an agent-agnostic manner.

\subsection{\label{sec:agent}Rule-Following Navigation Policy}
With a well-defined set of traffic rules, agents can follow them to achieve navigation while avoiding congestion. The real-world agent navigation unfolds in two distinct stages. In the initial stage, such as a scenario involving a vehicle driver, a routing algorithm is engaged—examples include popular tools like Google Maps—to chart a course toward the desired destination. However, these routing algorithms primarily rely on a coarse estimate of travel time and do not deeply incorporate knowledge of traffic rules into their decision-making process. Subsequently, when the driver embarks on the selected route, the traffic rules take center stage. The driver's journey is guided and informed by these regulations, ensuring smooth and unimpeded passage along the chosen route.

Our approach emulates this process through a two-stage navigation policy, denoted as $v_i^t=\pi(x_i^t,g_i)$. This policy takes the agent's present location $x_i^t$ and its designated goal position $g_i$ as inputs and subsequently generates the rule-following, goal-directed velocity $v_i^t$. 

During the first stage, our policy capitalizes on a visibility graph algorithm to ascertain the shortest trajectory connecting $x_i^t$ and $g_i$ (\prettyref{fig:pipeline}d). The visibility graph precomputes all pairs of vertices that are visible to each other without being blocked by other parts of the environment. The computation of such graph has been well-studied and we use the efficient algorithm proposed in~\cite{lozano1979algorithm}. Subsequent to this precomputation, a single visibility query is conducted per agent, establishing connections between $x_i^t$ and all visible graph nodes. A final shortest path query on the graph results in the sought-after trajectory. We denote this procedure as the shortest path function $\bar{v}_i^t=\SP(x_i^t,g_i)$, giving the tentative velocity that propels the agent along the optimal path.

Following the initial stage, we utilize our GRNN to modulate $\bar{v}_i^t$, transforming it into the rule-compliant velocity $v_i^t$. Given that traffic rules are inherently localized, we find it sufficient to query the nearest block within which our agent resides, as is indexed by its center position:
\begin{align}
\label{eq:indexing}
\ID(x_i^t)=\lfloor{x_i^t}\rfloor+0.5.
\end{align}
Subsequently, we enact the velocity modulation process via another MLP, define as the Rule-Following Unit (RFU):
\begin{align}
v_i^t=\TRU\left(h_{\ID(x_i^t)}^K,x_i^t-\ID(x_i^t),\bar{v}_i^t\right),
\end{align}
This MLP incorporates three crucial parameters: the hidden state $h_{\ID(x_i^t)}^K$ of the underlying block, encoding the local traffic rule; the relative position between the agent and the block's center $x_i^t-\ID(x_i^t)$; and the tentative velocity $\bar{v}_i^t$. As an optional measure to ensure training stability and maintain runtime performance, we have the option to constrain the velocities $\bar{v}_i^t$ and $v_i^t$ using the maximum agent velocity $v_\MAX$ via the clamping function: $v=v/\max\{1,(\|v\|+\epsilon)/v_\MAX \}$ where $\epsilon$ is a small regularization constant.

\revise{In real-world scenarios, agents are tasked with route selection, e.g., using Google Maps, followed by adherence to traffic rules. This paradigm underscores that the function $\pi(\bullet)$ is executed on the agent part, with the environment only storing the $h_{ij}^K$. However, in the context of a robotic system, a more nuanced distribution of responsibilities becomes feasible. For agents with limited computational capabilities, the agent only needs to store $x_i^t$ and $g_i$, and queries the environmental nodes for the desired $v_i^t$. For agents endowed with slightly more computational resources, agents can independently determine routes by running the $\SP(\bullet)$ function and command the environmental nodes to execute the $\TRU(\bullet)$ function. Significantly, regardless of the deployed setting, these agents operate in a fully decentralized manner, eliminating the need for inter-agent communication. Essentially, each agent is only required to identify the block and query the RFU.}

\subsection{\label{sec:train}Policy Optimization}
Harnessing the intrinsic locality of traffic rules, our policy is defined by three compact networks: $\HH$, $\EConv$, and $\TRU$. Nonetheless, optimizing this policy presents a formidable challenge due to the inherently non-smooth and stochastic nature of the navigation process. Due to the involvement of the non-differentiable block indexing operator (\prettyref{eq:indexing}), conventional model-based differentiable policy optimization techniques such as~\cite{ye2023differentiable} cannot be applied. Further, we consider intricate scenarios where agents can enter or exit the environment at arbitrary time points. Therefore, the collective state of agents undergoes dimension changes, deviating from the premises underpinning standard RL algorithms~\cite{vinitsky2018benchmarks}. Finally, agents can undertake detours to follow traffic rules and avoid local collisions, making it difficult to assess task accomplishment and gauge the extent of congestion. These challenges are well-recognized in the realm of RL in the context of sparse rewards~\cite{hare2019dealing}.

To confront these issues, we introduce two training approaches. When groundtruth traffic rules, denoted as $\pi^\star$, are known to the user, we adopt the IL paradigm~\cite{ross2011reduction} to optimize our policy by emulating the expert's behavior. In this vein, we assemble a dataset of simulation scenarios, collectively denoted as $\mathcal{S}={S_i}$. Each scenario, $S_i$, takes the form of a tuple: $S_i=<\FREE,\SEED(\bullet),\DELETE(\bullet)>$. Here, $\FREE$ represents the environment's freespace geometry. The function $\{x_i^t,g_i\}=\SEED(x_{1,2,\cdots}^t)$ generates new agents to enter the environment with respective goal positions, based on the current agent configuration. Similarly, $\{x_i^t\}=\DELETE(x_{1,2,\cdots}^t)$ determines which subset of agents have reached their goals and should exit the environment. During each iteration of IL training, we simulate navigations for a randomly selected scenario, $S_i\in\mathcal{S}$ over a span of $T$ timesteps. This simulation process accumulates a collection of agent transition tuples to form the dataset $\mathcal{D}=\{<x_i^t,g_i,\pi(x_i^t,g_i),\pi^\star(x_i^t,g_i)>\}$. IL proceeds by minimizing the following expert discrepancy:
\begin{align}
\mathcal{L}_\text{IL}(S_i,\theta)=\frac{1}{|\mathcal{D}|}\sum_{<x_i^t,g_i>\in\mathcal{D}}|\pi(x_i^t,g_i)-\pi^\star(x_i^t,g_i)|^2.
\end{align}
Here, $x_i^t$ and $g_i$ remain fixed variables, so the objective is differentiable under appropriate parameterization. 

When an expert rule is not available, we train our policy using evolutionary RL~\cite{salimans2017evolution}. Unlike conventional RL, this methodology bypasses the use of state-dependent rewards. Instead, it operates based on the assumption of a scenario-dependent sparse reward, denoted as $R(S_i,\theta)$, where $\theta$ corresponds to our learnable parameters, and aims to maximize the expected reward:
\begin{align}
\mathcal{L}_\text{RL}(S_i,\theta)=\mathbb{E}_{\epsilon\in\mathcal{N}(0,I)}[R(S_i,\theta+\epsilon)],
\end{align}
which essentially smoothens the reward signal. This technique iteratively updates $\theta$ using a stochastic gradient estimator of the following form:
\begin{equation}
\begin{aligned}
&\nabla_\theta \mathcal{L}_\text{RL}(S_i,\theta)
=\mathbb{E}_{\epsilon\in\mathcal{N}(0,I)}[R(S_i,\theta+\epsilon)\epsilon]/\sigma\\
\approx& \frac{1}{n\sigma}\sum_{k=1}^n R(S_i,\theta+\epsilon_k)\epsilon_k,
\end{aligned}
\end{equation}
where $\sigma$ denotes the estimated standard deviation and the estimator follows from log-likelihood technique and random sampling~\cite{wierstra2014natural}. To enhance performance, we adopt fitness shaping as detailed in~\cite{wierstra2014natural} and use the following alternative approximation:
\begin{equation}
\begin{aligned}
&\nabla_\theta \mathcal{L}_\text{RL}(S_i,\theta) \approx \frac{1}{\sigma}\sum_{k=1}^n u_k\epsilon_{k},\\
&u_k\triangleq\frac{\max \left(0, \log \left(\frac{n}{2}+1\right)-\log (k)\right)}{\sum_{j=1}^n \max \left(0, \log \left(\frac{n}{2}+1\right)-\log (j)\right)}-\frac{1}{n},
\end{aligned}
\end{equation}
where \( \{ R(S_i, \theta + \epsilon_{k}) \}_{k=1}^{n} \) is the descending ordered sequence derived from \( \{ R(S_i, \theta + \epsilon_k) \}_{k=1}^{n} \). The fitness shaping is known to improve the convergence as recommended in~\cite{salimans2017evolution}. Our overall algorithm follows a sequential routine, iteratively sampling of a batch of scenarios, executing simulations to assess $R(S_i,\theta)$, and forming the gradient estimator to update $\theta$.

The efficacy of the optimized policy hinges significantly on the reward signal $R(S_i,\theta)$. We discern that a well-constructed set of traffic rules should universally ensure freedom from congestion. That is, an agent can traverse from any starting point to any goal position without encountering congestion. Inspired by this insight, we tailor our reward to reflect the worst-case fraction of travel across all agents. In essence, we begin by identifying the set of agents that emerge during the simulation of $S_i$, denoted as $\mathcal{A}(S_i)={<x_i,g_i>}$. For each of these agents, we gauge their fraction of travel through the metric: $\FOC(x_i,g_i)=1-\SD(x_i^T,g_i)/\SD(x_i^0,g_i)$, where $\SD(\bullet)$ signifies the shortest-distance function computed using the visibility graph, analogous to $\SP(\bullet)$. Subsequently, we approximate the worst $\FOC(\bullet)$ over all agents, employing a soft-min function:
\begin{align}
R(S_i,\theta)=\frac{\sum_{<x_i,g_i>\in\mathcal{A}(S_i)}\FOC(x_i,g_i)e^{-\alpha\FOC(x_i,g_i)}}{\sum_{<x_i,g_i>\in\mathcal{A}(S_i)}e^{-\alpha\FOC(x_i,g_i)}},
\end{align}
where $\alpha$ is a softness-controlling parameter. To ensure training stability, we further incorporate curriculum learning that interleaves the increase of $\alpha$ with the update of $\theta$. The complete workflow of our evolutionary RL is outlined in \prettyref{alg:training}.

\begin{algorithm}
\caption{\label{alg:training} Evolutionary RL}
\begin{algorithmic}[1]
\Require{A set of scenarios $\mathcal{S}$, initial $\theta,\alpha$, parameter $B,\eta$}
\Ensure{The universally congestion-free $\pi(\bullet)$}
\While{$\theta$ not converged}
\State $\mathcal{D}\gets\emptyset, \theta_0\gets\theta$, sample $S_i\in\mathcal{S}$\Comment{Batch size $=1$}
\For{$i=1,\cdots,B$}
\State $\epsilon\in\mathcal{N}(0,I)$, $\theta\gets\theta_0+\epsilon$, $\{x_i^0,g_i\}\gets\emptyset$
\LineComment{Simulate scenario}
\For{$t=0,\cdots,T-1$}
\State Emit $\{x_i^t,g_i\}\gets\{x_i^t,g_i\}\cup\SEED(x_{1,2,\cdots}^t)$
\State Delete $\{x_i^t,g_i\}\gets\{x_i^t,g_i\}-\DELETE(x_{1,2,\cdots}^t)$
\LineComment{Local navigation algorithm (\NAVI)~\cite{van2008reciprocal,karamouzas2017implicit}}
\State Simulate $\{x_i^{t+1},g_i\}\gets\NAVI(\{x_i^t,g_i\})$
\EndFor
\State Collect $\mathcal{A}(S_i)$ and $\mathcal{D}\gets\mathcal{D}\cup\{<R(S_i,\theta),\epsilon>\}$
\EndFor
\State Evaluate $\nabla_\theta\mathcal{L}_\text{RL}(S_i,\theta)$ using $\mathcal{D}$
\State Update $\theta\gets\theta_0+\eta\nabla_\theta\mathcal{L}_\text{RL}$
\State Optionally increase $\alpha$
\EndWhile
\State Return $\theta$ and $\pi(\bullet)$
\end{algorithmic}
\end{algorithm}

%% file: results.tex
\section{\label{sec:results}Evaluation}
Our method is realized through a robust implementation using C++ and PyTorch. We employ C++ to implement the implicit crowd algorithm~\cite{karamouzas2017implicit}, serving as our foundational local navigation framework. Our simulations run in parallel across multiple scenarios. For uniformity and consistency across all our experiments, we adopt a standardized network architecture. Our policy is parameterized through the integration of three compact MLPs. The function $\EConv(\bullet)$ is parameterized as: FC32$\to$ReLU$\to$FC32$\to$ReLU$\to$FC32. The function $\HH(\bullet)$ is parameterized as: FC32$\to$ReLU$\to$FC32$\to$ReLU and we set $|h_{ij}|=32$. Finally, the function $\TRU(\bullet)$ is parameterized as: FC32$\to$ReLU$\to$FC32$\to$ReLU$\to$FC16$\to$ReLU$\to$FC2. For IL training, we set $T=500$  during each round of IL and run at least $1000$ iterations of Adam to optimize $\mathcal{L}_\text{IL}$ with a learning rate of $10^{-4}$. 
%Moreover, we finish the running round after updating parameters for $8000$ iterations or the following threshold is satisfied:
% \begin{align*}
% \frac{180}{\pi}\cdot\max_{<x_i^t,g_i>\in\mathcal{D}}
% \arg \cos \frac{\langle\pi(x_i^t,g_i),\pi^\star(x_i^t,g_i)\rangle}{\|\pi(x_i^t,g_i)\|\cdot\|\pi^\star(x_i^t,g_i)\|}\leq 40.
% \end{align*}
For RL training, we  set $T=500$, $\sigma=0.02$, $B=10$, the initial $\alpha=0$, which is then reduced by $0.01$ every $2000$ iterations. Note that we set $\eta=2\times10^{-4}$ for the first $5000$ iterations and $\eta=2\times10^{-5}$ for the rest. All experiments are performed on a single desktop machine with an AMD EPYC 7K62 CPU and we perform all computations on CPU, on which an IL training takes $48$ hours and an RL training takes one week.

\subsection{Scenario Dataset}
Our method requires two datasets: a training scenario dataset denoted as $\mathcal{S}$ and a testing scenario dataset referred to as $\mathcal{S}'$, with $\mathcal{S}'$ specifically comprising scenarios that have not been encountered during training. We set $|\mathcal{S}|$ to be $85$ and $120$ for the training dataset of RL and IL, respectively; We set $|\mathcal{S}'|=30$ for the testing dataset of both IL and RL. To enable dynamic navigation within each scenario $S_i$, we further manually create a collection of starting and goal positions. During each time step, our seeding function $\SEED(\bullet)$ continually generates new agents, originating from these predefined starting positions as long as the newly generated agents do not collide with any existing ones. In addition, we set the maximum number of agents for each scenario to be $40$ in the experiments. Conversely, our deleting function $\DELETE(\bullet)$ removes agents that have successfully reached the same block as their respective goal positions reside.

\subsection{Performance}
In our evaluation, we conduct a comparative analysis of our method against a baseline approach that relies solely on the shortest path without any rule-following modulation. This baseline policy, denoted as $\pi^b$, is achieved by setting $v_i^t=\bar{v}_i^t$. We gauge the performance of these policies using two fundamental metrics: the average fraction of travel $R_0$ and the worst fraction of travel $R_\infty$ defined as follows:
\begin{align*}
R_0=\frac{1}{|\mathcal{S}'|}\sum_{S_i\in\mathcal{S}'}R_{\alpha=0}(S_i,\theta)\quad
R_\infty=\frac{1}{|\mathcal{S}'|}\sum_{S_i\in\mathcal{S}'}R_{\alpha=\infty}(S_i,\theta).
\end{align*}

\begin{figure}[ht]
\centering
\includegraphics[width=.95\linewidth]{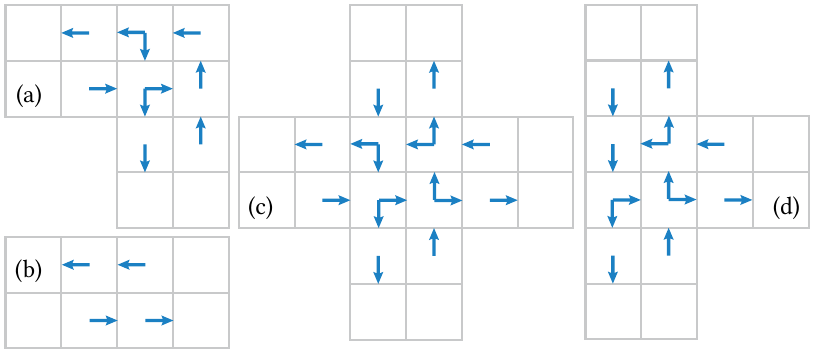}
\caption{\label{fig:expert}The set of allowed moving directions $\mathcal{V}_{ij}$ for 4 types of blocks used by our expert policy. We do not label the $\mathcal{V}_{ij}$ for boundary blocks, which depend on the boundary conditions therein.}
\end{figure}
\begin{figure}[ht]
\centering
\includegraphics[width=.95\linewidth]{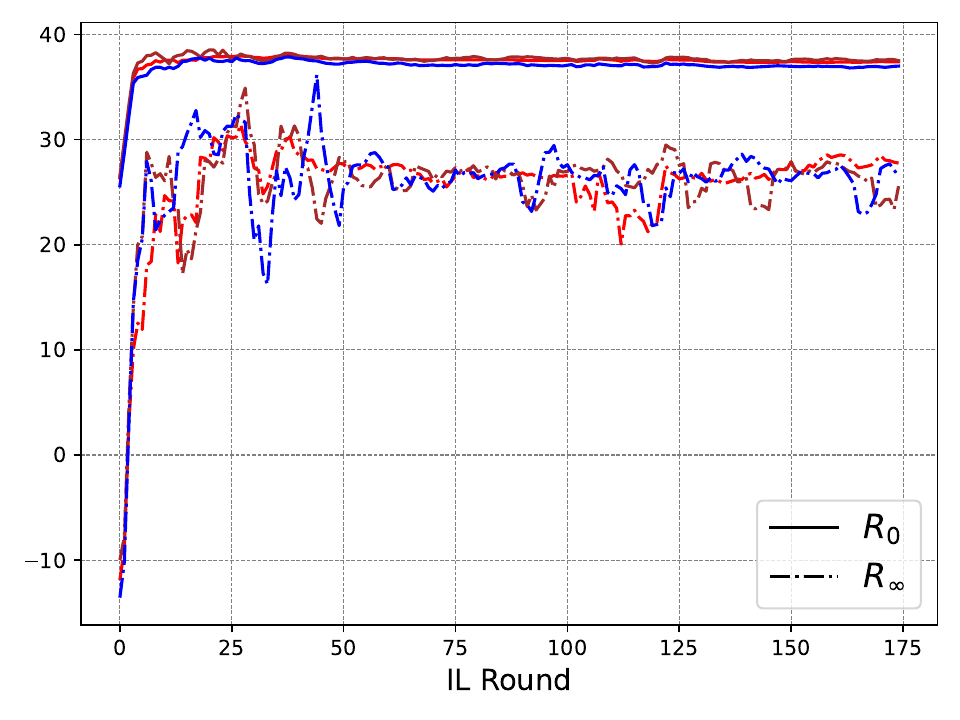}
\caption{\label{fig:ILConv}\revise{The convergence history in terms of $R_0$ and $R_\infty$ plotted against the index of IL round. We superimpose the history of 3 runs (red/blue/dark red) to highlight the consistency of our method.}}
\end{figure}

Our initial evaluation focuses on the performance of IL-trained policies, which necessitates a set of groundtruth traffic rules. To this end, we craft a set of traffic rules emulating real-world road networks. Each lane is designed to accommodate travel in only one specific direction. Furthermore, we incorporate the conventions of real-world roundabouts at road intersections, wherein all vehicles are constrained to move in a counter-clockwise direction. This set of rules yields a predefined collection of allowable moving directions, denoted as $\mathcal{V}_{ij}$ for the $ij$-th block. A visual representation of these directions is provided in~\prettyref{fig:expert}. Our expert policy then determines the modulated velocity as one of the allowed moving directions that aligns best with $\bar{v}_i^t$ as defined through the following expression:
\begin{align*}
\pi^\star(x_i^t,g_i)=\argmax{v\in\mathcal{V}_I(x_i^t)}<v,\bar{v}_i^t>.
\end{align*}
To track the convergence of our IL training process, we present a convergence history plot in~\prettyref{fig:ILConv}. Notably, this plot illustrates that IL rapidly converges often requiring as few as $50$ rounds. Following training, we summarize the performance of all three policies in~\prettyref{table:IRLPerf}. It is evident that there exists a minor performance gap in terms of $R_0$ between $\pi^\text{IL}$ and $\pi^\star$, which provides confirmation that our meticulously parameterized neural policy possesses the capability to effectively capture and express a subset of real-world traffic rules. However, $\pi^{\text{IL}}$ is vulnerable to variations in $R_{\infty}$ when evaluated in new scenarios. This limitation is mitigated by $\pi^{\text{RL}}$, which does not strictly adhere to $\pi^\star$ and thus has greater flexibility in adapting to unfamiliar situations. Furthermore, both $\pi^\text{IL}$ and $\pi^\star$ exhibit notably superior performance compared to $\pi^b$, demonstrating the remarkable effectiveness of incorporating real-world traffic rules in alleviating congestion-related issues.
\begin{table}[ht]
\centering
\caption{\label{table:IRLPerf}Testing score of baseline $\pi^b$, RL-trained $\pi^\text{RL}$, IL-trained $\pi^\text{IL}$, and the expert $\pi^\star$ in terms of $R_0$ and $R_\infty$. We run all the test scenarios $10$ times and report the average$\pm$deviation.}
\begin{tabular}{ccccc}
\toprule
& $\pi^b$ & $\pi^\text{IL}$ & $\pi^\text{RL}$ & $\pi^\star$\\
\midrule
$R_0$ & $25.35\pm3.35$  & $33.97\pm1.98$ & $38.36\pm 0.52$ & $38.45\pm 0.07$\\
$R_\infty$ & $5.13\pm 2.33$  & $9.31\pm6.75$ & $37.78\pm 0.18$ & $38.06\pm 0.10$\\
\bottomrule
\end{tabular}
\end{table}
\begin{figure}[ht]
\centering
\includegraphics[width=.95\linewidth]{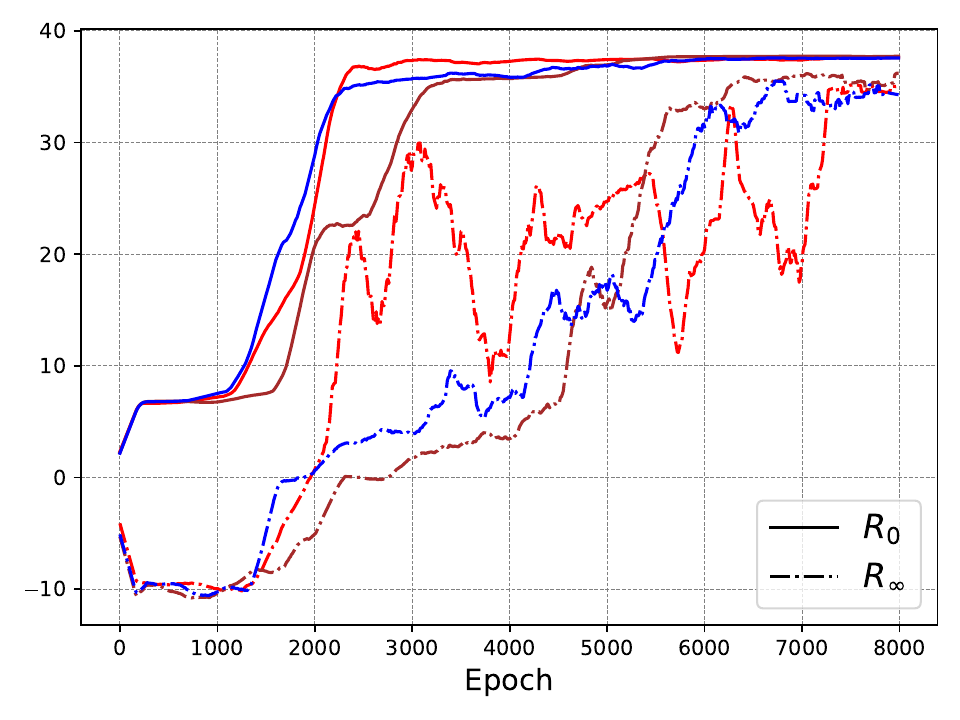}
\caption{\label{fig:RLConv}\revise{The convergence history in terms of $R_0$ and $R_\infty$ plotted against the index of RL iteration. We superimpose the history of 3 runs (red/blue/dark red) to highlight the consistency of our method.}}
\end{figure}

\input{illus.tex}
Moving on, our evaluation proceeds to the RL-trained policy. Again we plot the convergence history in~\prettyref{fig:ILConv} and summarize its performance in the second column of~\prettyref{table:IRLPerf}. Some instability is observed during the training process as we use smaller values of $\alpha$ leading to progressively non-smooth reward signals. In addition, this behavior can be attributed to the small value of $B=10$ causing imprecise approximation of the policy gradient. Conversely, opting for larger values of $B$ can significantly escalate the training cost. Upon completion of training, we observe a distinguished improvement in terms of $R_\infty$ compared to $\pi^\text{IL}$. This result is consistent with our expectations, as RL autonomously searches for optimal traffic rules without requiring expert guidance, thereby enhancing robustness. As anticipated, both the IL and RL-trained policies substantially outperform $\pi^b$. This underscores not only the expressiveness of our method but also its capacity to autonomously discover suitable traffic rules, guided solely by congestion-resolving reward signals. A visual comparison is given in~\prettyref{fig:scenarios}.

\begin{figure}[ht]
\centering
\includegraphics[width=.95\linewidth]{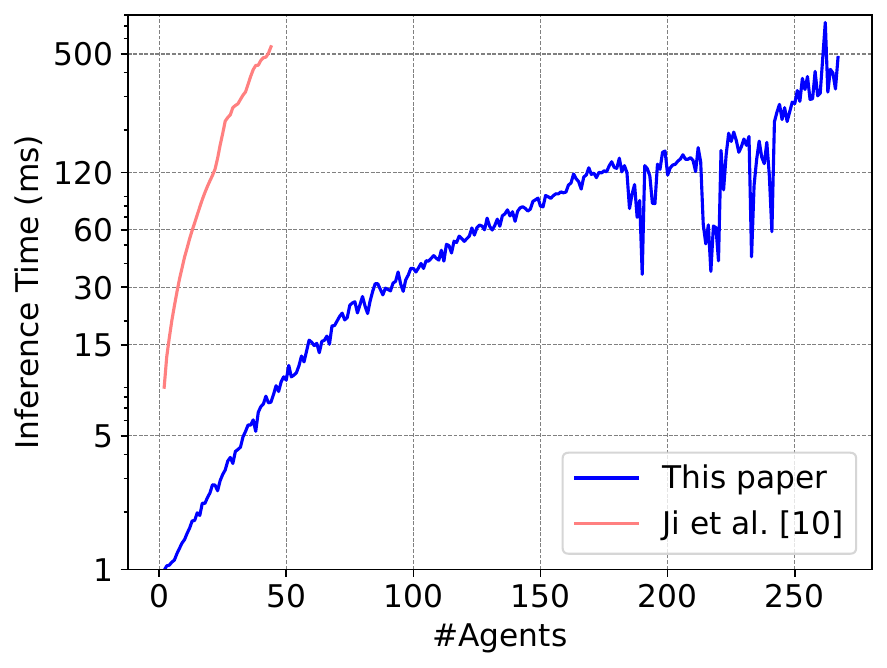}
\caption{\label{fig:FPS}\revise{The runtime total inference cost profiled in milliseconds, which is averaged over the 10 runs on test scenarios.}}
\end{figure}
\revise{Ultimately, we evaluate the runtime inference costs of our approach and the agent-centric method~\cite{ji2021decentralized} by comparing them in~\prettyref{fig:FPS}. Both methodologies utilize GNNs to parameterize the policy. For this comparison, we pick a random instance from our scenario with a variable number of agents. We then run the GNN inference of the agent-centric method on the agent network, using the same parameter setting as~\cite{ji2021decentralized}. Remarkably, our architecture conducts GRNN inference only once, leading to significantly enhanced performance. As depicted in~\prettyref{fig:FPS}, we attain a peak performance of $400$ milliseconds while managing a cohort of $240$ agents.}

%% file: illus.tex
\begin{comment}
\begin{figure*}[ht]
\centering
%\vspace{-10px}
\setlength{\tabcolsep}{1px}
\begin{tabular}{cccc}
\includegraphics[width=.28\linewidth]{figs/comparison/env_1_SP.pdf}&
\includegraphics[width=.225\linewidth]{figs/comparison/env_2_SP.pdf}&
\includegraphics[width=.20\linewidth]{figs/comparison/env_3_SP.pdf}&
\includegraphics[width=.23\linewidth]{figs/comparison/env_4_SP.pdf}\\
\includegraphics[width=.28\linewidth]{figs/comparison/env_1_RL.pdf}&
\includegraphics[width=.225\linewidth]{figs/comparison/env_2_RL.pdf}&
\includegraphics[width=.20\linewidth]{figs/comparison/env_3_RL.pdf}&
\includegraphics[width=.23\linewidth]{figs/comparison/env_4_RL.pdf}\\
\end{tabular}
%\vspace{-5px}
\caption{\label{fig:scenarios}
We show congested scenarios caused by using the baseline $\pi^b$ (top) and the corresponding congestion-free scenarios achieved by our $\pi^\text{RL}$ (bottom). The left two scenarios are from training scenarios and the right two are from testing scenarios. Agents are colored according to their group, where each group has distinctive initial and goal positions.}
\vspace{-10px}
\end{figure*}
\end{comment}

\begin{figure}[ht]
\centering
\includegraphics[width=.45\linewidth]{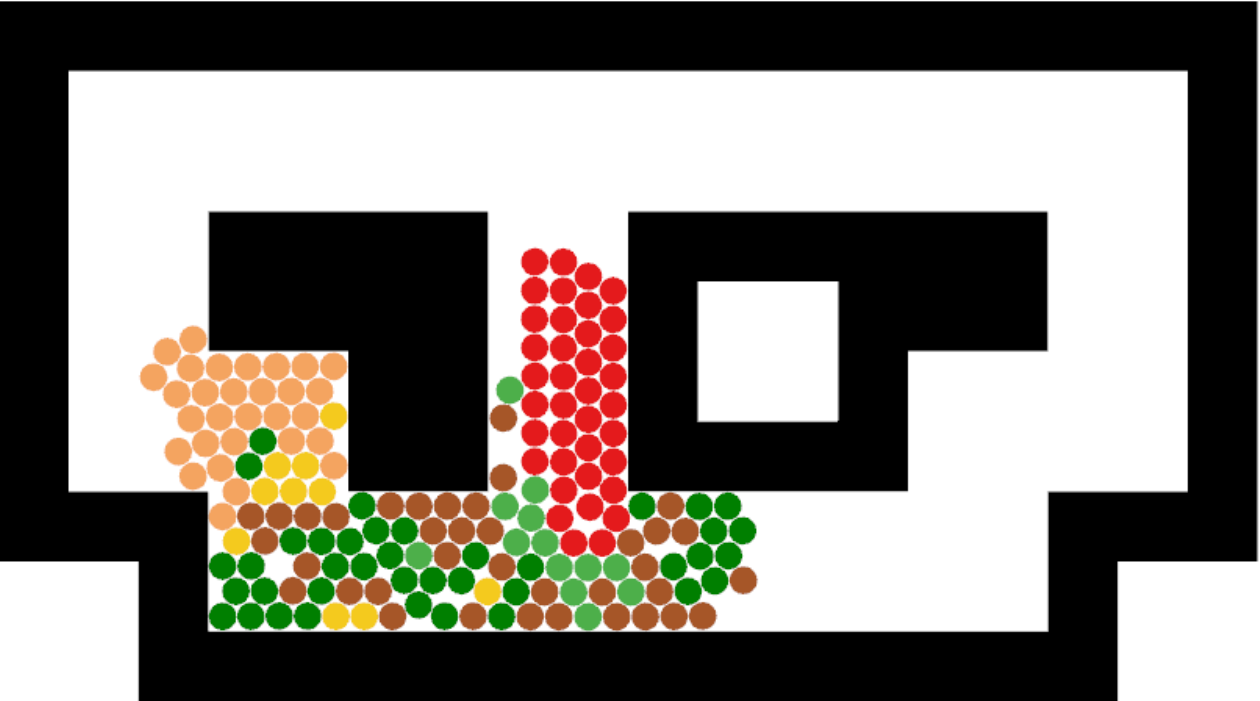}
\includegraphics[width=.45\linewidth]{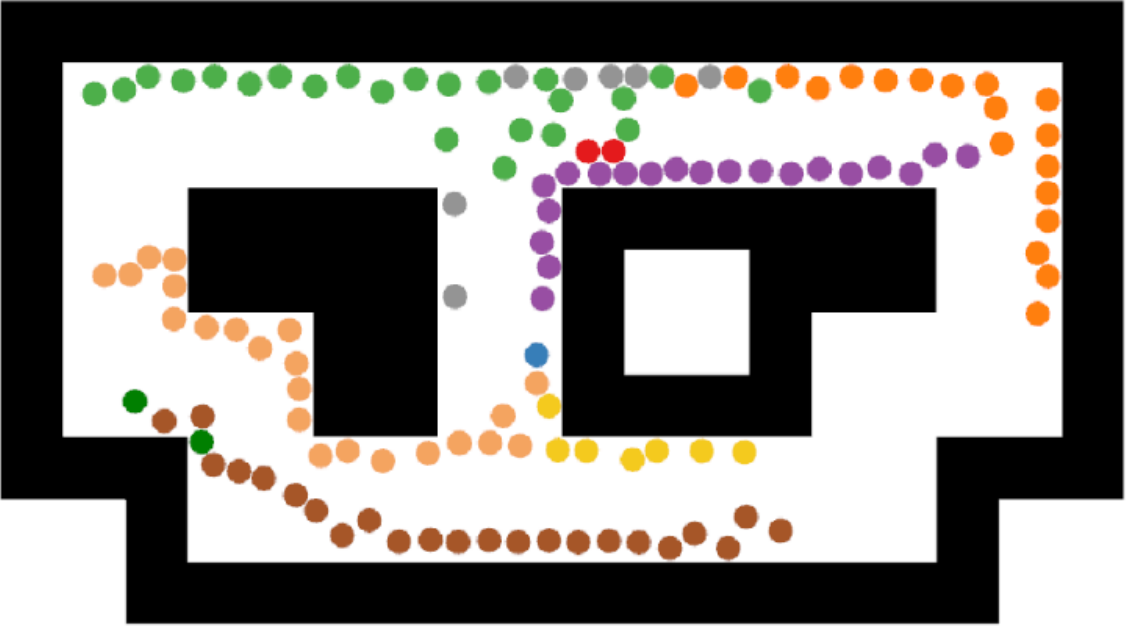}
\caption{\label{fig:scenarios}
We show congested scenarios caused by using the baseline $\pi^b$ (left) and the corresponding congestion-free scenarios achieved by our $\pi^\text{RL}$ (right). Agents are colored according to their group.}
%\vspace{-5px}
\end{figure}

%% file: conclusion.tex
\section{\label{sec:conclusion}\revise{Conclusion \& Future Works}}
We introduce an environment-centric architecture for neural navigation policies, designed to replicate real-world traffic networks. Our approach demonstrates the feasibility of training such a policy using IL guided by expert traffic rules or RL with congestion-resolving reward signals. \revise{Our approach eliminates the need for runtime inter-agent communication, resulting in significantly reduced computational burden compared to agent-centric navigation policies.}

This research opens up a plethora of promising avenues for future exploration, such as hardware deployment. One notable aspect to consider is the modeling of time-dependent traffic rules, such as those associated with traffic lights. Furthermore, our current method exclusively accounts for homogeneous traffic rules, yet in the real world, rule variations are contingent upon the context of the environment, encompassing scenarios like parking lots and highway checkpoints. Extending our approach to encompass these more nuanced factors represents an intriguing area for future research endeavors. \revise{Finally, our training procedure is time-consuming and CPU-bound. This is attributed to our collision-free agent simulator~\prettyref{eq:simulator}, which involves solving a joint optimization problem encompassing all agents. We plan to utilize parallel computations to speed up this procedure and expedite the training process.}